\documentclass[eat,twocolumn]{jmlr}

\usepackage{longtable}

\usepackage{booktabs}
\usepackage[load-configurations=version-1]{siunitx} 

\theorembodyfont{\upshape}
\theoremheaderfont{\scshape}
\theorempostheader{:}
\theoremsep{\newline}

\jmlrvolume{}
\firstpageno{1}

\jmlryear{2022}
\jmlrworkshop{Machine Learning for Health (ML4H) 2022}

\usepackage{amssymb,amsmath,mathtools,bbm}
\usepackage{algorithm,algpseudocode}
\usepackage{graphicx}
\usepackage{blindtext}
\usepackage{capt-of}
\usepackage{booktabs}
\usepackage{varwidth}
\newsavebox\tmpbox

\title[Decision-Aware Learning for Optimizing Health Supply Chains]{Decision-Aware Learning \\ for Optimizing Health Supply Chains}

\author{\Name{Tsai-Hsuan Chung} \Email{angelchg@wharton.upenn.edu}\\
\addr University of Pennsylvania
\AND
\Name{Vahid Rostami} \Email{vahid@macro-eyes.com}\\
\addr Macro-Eyes
\AND
\Name{Hamsa Bastani} \Email{hamsab@wharton.upenn.edu}\\
\addr University of Pennsylvania
\AND
\Name{Osbert Bastani} \Email{obastani@seas.upenn.edu}\\
\addr University of Pennsylvania
}

\begin{document}

\maketitle

\begin{abstract}
We study the problem of allocating limited supply of medical resources in developing countries, in particular, Sierra Leone. We address this problem by combining machine learning (to predict demand) with optimization (to optimize allocations). A key challenge is the need to align the loss function used to train the machine learning model with the decision loss associated with the downstream optimization problem. Traditional solutions have limited flexibility in the model architecture and scale poorly to large datasets. We propose a decision-aware learning algorithm that uses a novel Taylor expansion of the optimal decision loss to derive the machine learning loss. Importantly, our approach only requires a simple re-weighting of the training data, ensuring it is both flexible and scalable, e.g., we incorporate it into a random forest trained using a multitask learning framework. We apply our framework to optimize the distribution of essential medicines in collaboration with policymakers in Sierra Leone; highly uncertain demand and limited budgets currently result in excessive unmet demand. Out-of-sample results demonstrate that our end-to-end approach can significantly reduce unmet demand across 1040 health facilities throughout Sierra Leone.
\end{abstract}

\section{Introduction}

A major challenge for healthcare administration in developing nations is the ability to effectively allocate limited resources across health facilities. We study an instance of this problem for Sierra Leone, where every three months, the national government must allocate limited supplies of many different essential medicines to over a thousand health facilities based on the expected demand at those facilities. Failure to allocate sufficient quantities to facilities---known as \emph{unmet demand}---can have significant health consequences.

\begin{figure}
\centering
\includegraphics[width=0.9\columnwidth]{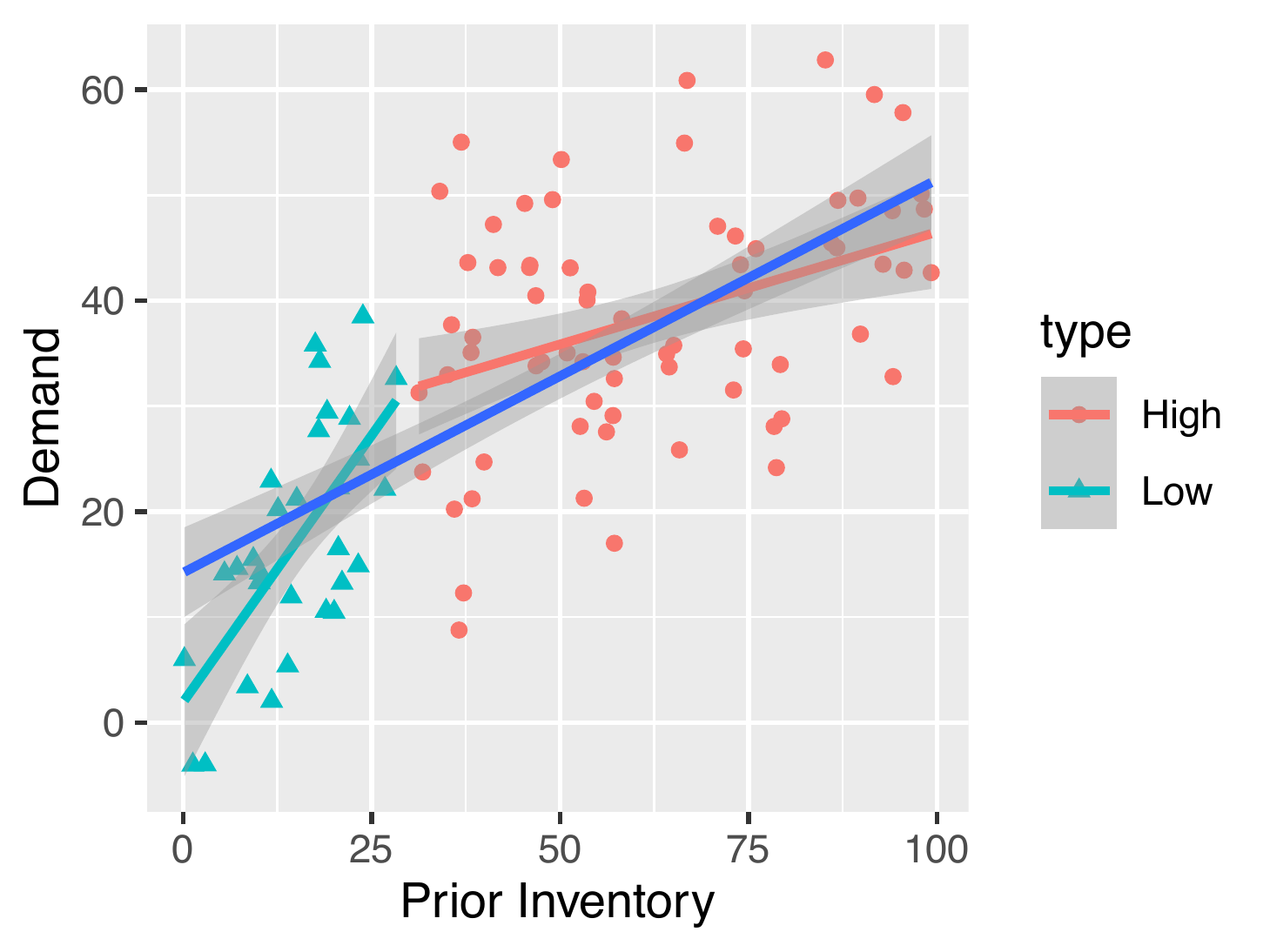}
\caption{Intuitive example}
\label{fig:example}
\end{figure}

Combining machine learning and optimization is a powerful approach to addressing such decision-making problems. Machine learning is critical for making accurate predictions of the future demand for medicines; these forecasts can then be incorporated into downstream optimization algorithms to optimize allocations. The most na\"{i}ve instantiation of this strategy is called \emph{predict-then-optimize}, where the machine learning model is trained to predict demand using a typical objective such as mean-square error (MSE), completely independently of the downstream optimization problem.

However, this strategy can result in poor performance since it may not focus the capacity of the machine learning model on predictions that are actually relevant to making decisions---e.g., it may ``spend'' a lot of capacity on predicting demand for facilities that are unlikely to run out of supplies. For intuition, consider a stylized scenario where there are two kinds of facilities: well-stocked facilities, which do not require additional inventory, and poorly-stocked facilities, which require additional inventory.

For an intuitive example illustrating the issue with predict-then-optimize, consider the (synthetic) data in Figure~\ref{fig:example}. In this example, there are two kinds of facilities: wealthier ones with high prior inventory (red) and poorer ones with low prior inventory (blue). The high-inventory facilities do not need additional inventory, so they can be ignored by the optimization algorithm. They also exhibit different demand patterns: for high-inventory facilities, demand scales more slowly with prior inventory compared to low-inventory facilities---intuitively,
wealthier locations can keep arbitrary excess inventory, making their prior inventory less correlated with their demand.

Suppose that we use a linear model with prior inventory as a feature to predict demand. Because the problem is misspecified, the linear model will interpolate between the two kinds of facilities. Thus, even in the limit of infinite data, the model will make errors, thereby reducing the quality of the downstream optimization algorithm.

As a consequence, there has been significant interest in algorithms that incorporate the \emph{decision loss} into the prediction algorithm~\citep{kotary2021end, bertsimas2020predictive}. In other words, one might want to directly optimize the decision loss when training the machine learning model. One approach is to incorporate the decision loss into the learning algorithm---e.g., modifying the rule for splitting nodes in a decision tree to account for the decision loss~\citep{kallus2022stochastic}. However, this strategy requires a custom strategy for incorporating the decision loss into every new model family. A more general approach is to directly optimize the true decision loss when training the model---i.e.,
\begin{align}
\label{eqn:optloss}
\hat\theta=\operatorname*{\arg\max}_{\theta}\mathbb{E}_{p(x,\xi^*)}\left[\ell(\pi^*(f_\theta(x));\xi^*)\right],
\end{align}
where
\begin{align*}
\pi^*(\xi)&=\operatorname*{\arg\max}_a\ell(a;\xi) \nonumber
\end{align*}
where $x\in\mathcal{X}$ are the features for predicting demand (e.g., historical demand for the last few months), $\xi^*$ is the true demand, $f_\theta(x)$ is the demand predicted by model $f_\theta$ with parameters $\theta$, $\pi^*(\xi)$ is the optimal policy assuming the true demand is $\xi$, and $\ell(a;\xi^*)$ is the decision loss of allocation $a$ when the true demand is $\xi^*$. Algorithms for solving this optimization problem have been proposed in the linear setting~\citep{elmachtoub2022smart} as well as the more general setting of differentiable model families by taking gradients through the optimization problem~\citep{wilder2019melding, wang2020automatically}.

However, two critical shortcomings remain. First, several of these techniques are very specialized not only to specific model families, but also to a specific way of setting up the prediction problem. In particular, most existing approaches assume that there is a single prediction per optimization problem---e.g., we train a separate model $f_\theta^{(n)}$ to predict each component $\xi^*_n$ of the demand vector $\xi^*\in\mathbb{R}^N$ (where $N$ is the number of facilities). However, this strategy cannot adapt to more complex, modern data science pipelines. For example, in our allocation problem in Sierra Leone, training a separate predictive model for each health facility performs very poorly since there is very little historical data available for any given facility. Instead, we can substantially improve predictive performance by using multi-task learning---i.e., train a single model \emph{across} all facilities and health products in order to learn cross-facility, cross-product correlations.

Second, many of these techniques become computationally intractable as the problem becomes large, especially the more general approach based on taking gradients through the optimization problem~\citep{wilder2019melding, wang2020automatically}, since they require re-solving the optimization problem on every gradient step.

We propose a novel technique that is simultaneously significantly more computational tractable and flexible enough to handle a broad class of model families.

For instance, in our example in Figure~\ref{fig:example}, we can significantly improve performance simply by downweighting data from high-inventory facilities; since these facilities are ignored by the optimization algorithm, having high prediction accuracy does not help make better decisions. Then, the linear model would converge to the true model for the low-inventory facilities, enabling the optimization problem to make optimal allocations.

To this end, our strategy is to Taylor expand the optimal decision loss (\ref{eqn:optloss}), and show how this expansion can be interpreted as a reweighting of samples in the dataset; thus, our algorithm is ultimately implemented as training the underlying model with weights associated with each example.
Importantly, our approach can be straightforwardly integrated into complex data science since most machine learning models can use weighted training data. Furthermore, in our approach, the optimization problem only needs to be run \textit{once} to compute the weights, ensuring it is computationally efficient.

We evaluate our approach on historical demand data for essential medicines from Sierra Leone. Our experiments demonstrate that our approach outperforms the commonly-used predict-then-optimize approach; none of the existing decision-aware learning algorithms \citep{wilder2019melding, wang2020automatically, kallus2022stochastic} were able to scale to our dataset. We are currently working towards deploying our algorithm as a decision support tool for the Sierra Leone National Medical Supplies Agency.

\section{Decision Aware Inventory Allocation}

\textbf{Problem formulation.}
In the inventory allocation problem, the goal is to allocate a limited quantity of a single resource (e.g., a medicine) to facilities (e.g., hospitals, clinics, etc.) in a way that minimizes the \emph{expected shortfall} (or \emph{unmet demand})---i.e., the amount of demand from customers (e.g., patients) across facilities for which supply is unavailable. For simplicity, we assume that each customer only visits a single facility, and their request is either fulfilled or unfulfilled.

We assume that the amount to be allocated in any given quarter is a constant $a_{\text{max}}\in\mathbb{R}$, and our goal is to allocate this resource across $N\in\mathbb{N}$ facilities. Each facility $n\in[N]=$ has demand $\Xi_n$, where $\Xi\in\mathbb{R}^N$ is a real-valued random vector with distribution $\mathbb{P}_{\Xi}$. We denote our allocation decision by $a\in\mathbb{R}^N$, where $a_n$ is the allocation intended for facility $n$. Our objective is the \emph{expected shortfall}
\begin{align*}
\mathbb{E}_{\Xi}\left[\sum_{n\in[N]}\max\{\Xi_n-a_n,0\}\right],
\end{align*}
which measures the unmet demand on average across facilities and over $\Xi$.

\textbf{Optimization strategy.}
First, when $\Xi=\xi$ is constant, the optimal policy can be straightforwardly expressed as a linear program. To account for the randomness in $\Xi$, we use \emph{sample average approximation (SAA)}, which takes $K$ samples $\xi^{(k)}\sim\mathbb{P}_{\Xi}$ (for $k\in[K]$), and then optimizes the objective on average across these samples. The resulting optimization problem is
\begin{align}
\label{eqn:lp}
a^*=&\operatorname*{\arg\min}_{a\in\mathbb{R}^N}\frac{1}{K}\sum_{k=1}^K\sum_{n=1}^Nc_n^{(k)} \\
&\text{subj. to}\quad
c^{(k)}\ge \xi^{(k)}-a,
\quad
c\ge0, \\
&\qquad\qquad
\sum_{n=1}^Na_n\le a_{\text{max}}, \nonumber
\end{align}
where vector inequalities are elementwise. The first two constraints ensure $c_n^{(k)}=\max\{\xi_n^{(k)}-a_n,0\}$ (one of the constraints must bind to minimize the objective), and the last ensures the allocation budget. 

\textbf{Decision-aware loss.}
So far, we have assumed that the distribution of demand $\Xi$ is known, so we can draw samples from it to construct our objective. However, in our problem setting, the demand distribution is unknown, and we need to learn a model to predict it from data. To this end, we assume given a dataset $M=\{(x^{(m)},\xi^{*(m)})\}_{m\in[M]}$, where $x^{(m)}\in\mathbb{R}^d$ is a feature vector (e.g., demand during the last few periods) and $\xi^{*(m)}\in\mathbb{R}^N$ is the demand vector we are trying to predict.

Now, a na\"{i}ve stratey is to train $f_\theta$ to minimize a loss such as the mean-squared error---i.e., $\hat\theta=\arg\min_\theta\sum_{m=1}^M\|f_\theta(x)-\xi^{*(m)}\|^2$. However, this na\"{i}ve strategy does not account for how predictions account for the downstream optimization problem. We propose an approach that builds on the optimal decision loss in (\ref{eqn:optloss}).
Rather than directly try to optimize the decision loss optimal decision loss in (\ref{eqn:optloss}), which limits generality and can be computationally intractable, our approach focuses on upweighting data more relevant to the downstream optimization problem. To this end, we first approximate the decision loss by Taylor expanding it around $\hat\xi-\xi_0$ (where $\xi_0$ is the current prediction and $\hat\xi=f_\theta(x)$), yielding
\begin{align}
\label{eqn:taylor}
&\ell(\pi^*(\hat{\xi});\xi^*) \nonumber \\
&\approx
\ell(\pi^*(\xi_0);\xi^*) \nonumber \\
&+
\nabla_a\ell(\pi^*(\xi_0);\xi^*)^\top\nabla_\xi\pi^*(\xi_0)^\top(\hat{\xi}-\xi_0).
\end{align}
Since the first term is a constant, we can ignore it; then, we consider the loss
\begin{align*}
&\tilde{\ell}(\hat\xi;\xi^*) \\
&=\nabla_a\ell(\pi^*(\xi_0);\xi^*)^\top\nabla_\xi\pi^*(\xi_0)^\top(\hat\xi-\xi_0) \\
&=\nabla_a\ell(\pi^*(\xi_0);\xi^*)^\top\nabla_\xi\pi^*(\xi_0)^\top(\hat\xi-\xi^*)+\text{const}.
\end{align*}
Here, we have replaced $\xi_0$ with $\xi^*$; since both of these are constants, it does not affect the optimal solution, and we drop the constant term. With this replacement, we can upper bound the term by its absolute value to avoid ``overshooting'' $\xi^*$, resulting in a weighted absolute error loss:
\begin{align*}
\tilde{\ell}(\hat\xi;\xi^*)
=\sum_{n=1}^Nw_n\cdot(\hat\xi_n-\xi^*_n)
\le\sum_{n=1}^N|w_n||\hat\xi_n-\xi^*_n|,
\end{align*}
where
\begin{align*}
w_n=\left(\nabla_\xi\pi^*(\xi^*)\nabla_a\ell(\pi^*(\xi^*);\xi^*)\right)_n
\end{align*}
Our algorithm is to use this upper bound as the loss, which works with any standard machine learning algorithm that can take weighted examples. As a heuristic, we can replace the absolute error with the squared error. Finally, we train our random forest to minimize this loss on the training data:
\begin{align*}
\hat\theta=\operatorname*{\arg\min}_{\theta}\sum_{m=1}^M\sum_{n=1}^N|w_{m,n}||f_{\theta}(x^{(m)})_n-\xi^{*(m)}_n|.
\end{align*}

\textbf{Predict a demand distribution.}
Recall that we need a distribution over demand rather than a point estimate. We use a natural strategy---namely, we consider each tree in the random forest to be an i.i.d. sample from this distribution. In more detail, assuming the random forest $f_{\theta}$ consists of $K$ trees $f_{\theta}^{(1)},...,f_{\theta}^{(K)}$, we use samples $\xi_k=f_{\theta^{(k)}}(x)$ for $k\in[K]$ in SAA.

\textbf{Multitask learning.}
Rather than train a separate random forest to predict each component $\xi_n$ of the demand $\xi$, we use multitask learning to train a single model that can predict demand across all facilities and products. Specifically, we construct a feature vector $x_n\in\mathbb{R}^d$ for each facility, so the full set of features forms a matrix $x\in\mathbb{R}^{N\times d}$. Then, we train a model to predict $\xi^*_n$ from $x_n$---i.e., $f_{\theta}(x_n)\approx\xi^*_n$. We use this same strategy for both the initial random forest trained using the MSE loss as well as the final random forest trained using the decision aware loss.

\section{Experiments}

\begin{figure}[t]
\centering
\includegraphics[width=0.8\linewidth]{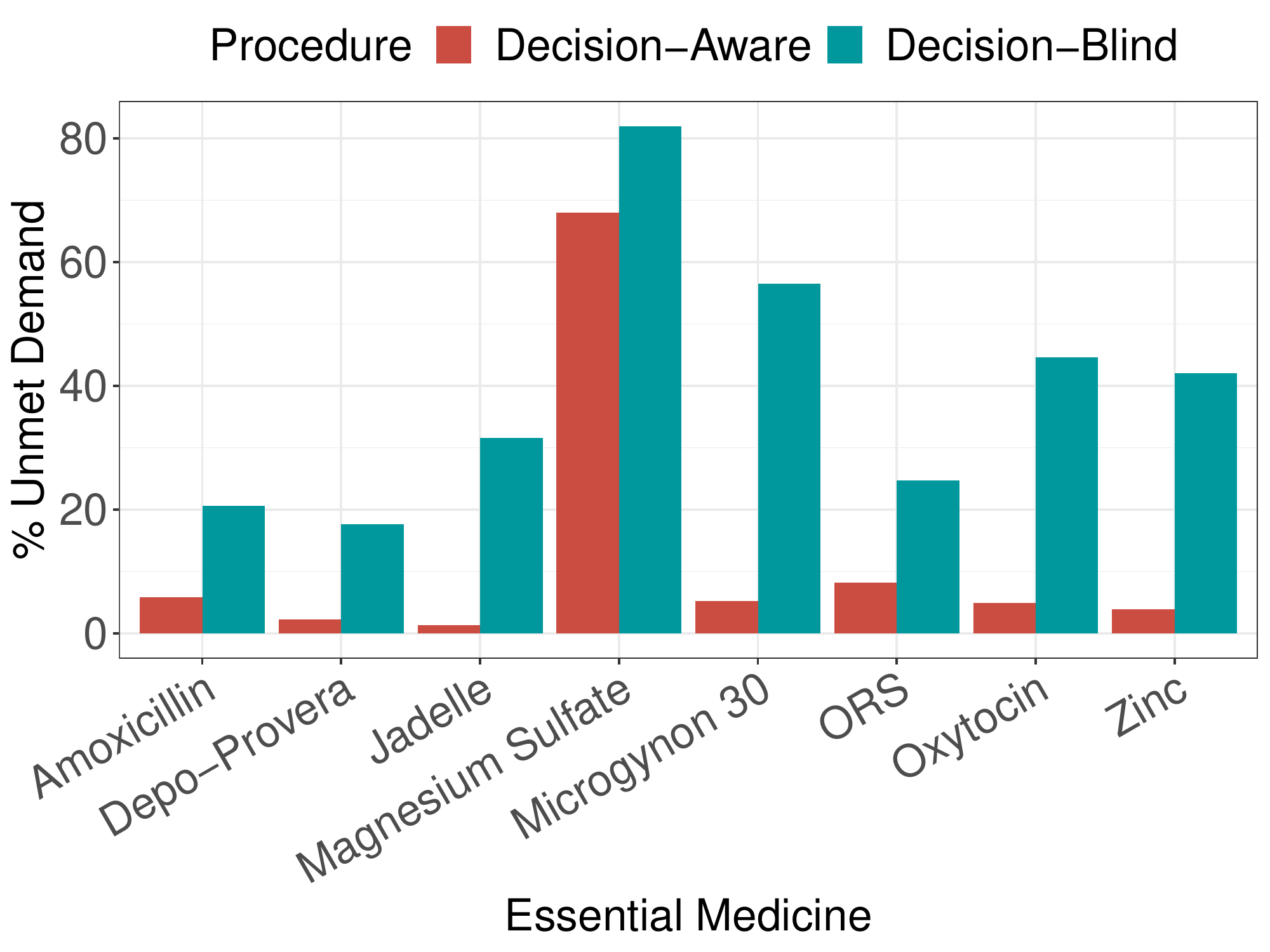} 
\caption{Comparison between the decision-blind predict-then-optimize approach and our decision-aware end-to-end approach. Both leverage a random forest model (trained with a MSE loss vs. approximate decision loss respectively) with the same downstream stochastic optimization problem.}
\label{fig:results}
\end{figure}

\textbf{Dataset.}
We use historical data sourced from District Health Information Software (DHIS2) provided by the Ministry of Health and Sanitation (MoHS) of Sierra Leone. This data includes the recorded demand of different essential medicines at 1040 medical facilities over the past 71 months (March 2019 to January 2022); we use 70 of these months for training and the final month for evaluation.
See the Appendix for further details on our predictive model.

We focus on the predictions and allocations for 8 medicines that were deemed most critical by policymakers in Sierra Leone---specifically, Amoxicillin, Magnesium Sulphate, Oxytocin, Zinc Sulphate, Oral Rehydration Salts, Depot Medroxyprogestrone Acetate, Ethinylestradiol \& Levonorgestrel (Microgynon 30), and Jadelle).

\textbf{Baselines.}
We compare to the popular decision-blind predict-then-optimize approach, which uses the MSE loss to train the random forest. We also tried existing decision-aware approaches on our dataset: decision-focused learning~\citep{wang2020automatically} and stochastic optimization random forests~\citep{kallus2022stochastic}. However, none of these scaled to our dataset. In particular, \citep{kallus2022stochastic} failed to terminate after running for two days on 5 threads running on 2$\times$ AMD EPYC 7402 Processor 24-core 2.80GHz CPUs. Similarly, the performance of~\citep{wang2020automatically} was very poor after running for several hours, likely because it only completed three epochs of training (recall that this approach requires solving the optimization problem at every gradient step).

\textbf{Results.}
We show results using both our approach and our baseline in Figure~\ref{fig:results}. We report the decision loss (i.e., the unmet demand) achieved for each of the 8 essential medicines as a percentage of total demand. As can be seen, our approach significantly outperforms the baseline on all products, by an order of magnitude in several cases. In addition, we also evaluated the robustness by looking at 10 facilities that have high allocation across products and another 10 facilities with low allocation across products. The results show that for higher allocation units, average unmet demand \% of our approach is 13\% compared with 18\% for decision-blind; for low allocation units, average unmet demand \% of our approach is 11\% compared with 17\% for decision-blind.
Our results demonstrate how our approach can bring the benefits of decision aware learning to realistic datasets and optimization problems by significantly improving downstream decision-making performance. In the Appendix, we also compare with the current practice in Sierra Leone and show a significant improvement. 

\section{Conclusion}

We have proposed a novel strategy for decision-aware learning that uses a Taylor expansion around the optimal solution to derive sample weights for training an arbitrary predictive model. Compared to prior approaches, our approach can straightforwardly be incorporated into many model architectures, and furthermore scales easily to large datasets. We have evaluated our approach on allocating essential medicines in Sierra Leone, demonstrating that our approach can produce as much as an order of magnitude improvement in performance compared to a na\"{i}ve predict-then-optimize strategy; furthermore, we have shown that existing decision-aware algorithms do not scale to this dataset. We believe that our work opens the door to leveraging decision-aware learning in a variety of realistic decision-making problems, including critical allocation and optimization problems in healthcare.

{\small
\bibliography{refs}
}

\newpage
\section*{Appendix}

\textbf{Observations.} We treat the demand for a (product, facility, month) tuple as a single observation in our data.

\textbf{Data Cleaning.} Health data, particularly from developing countries, can be messy with many missing or unreliable values. Therefore, we took several steps to ensure that our model was trained on reliable data. First, we excluded any observations where the Opening Balance + Quantity Received - Quantity Dispensed + Adjustment/Loss did not equal the Closing Balance. Second, we excluded observations where all the quantities on record were zero, since this suggests they used a default value. Finally, we excluded large outliers (e.g., likely reporting doses rather than vials). This results in 136,765 reliable training and 4920 reliable test observations.

\begin{figure}[t]
\centering
\includegraphics[width=1\linewidth]{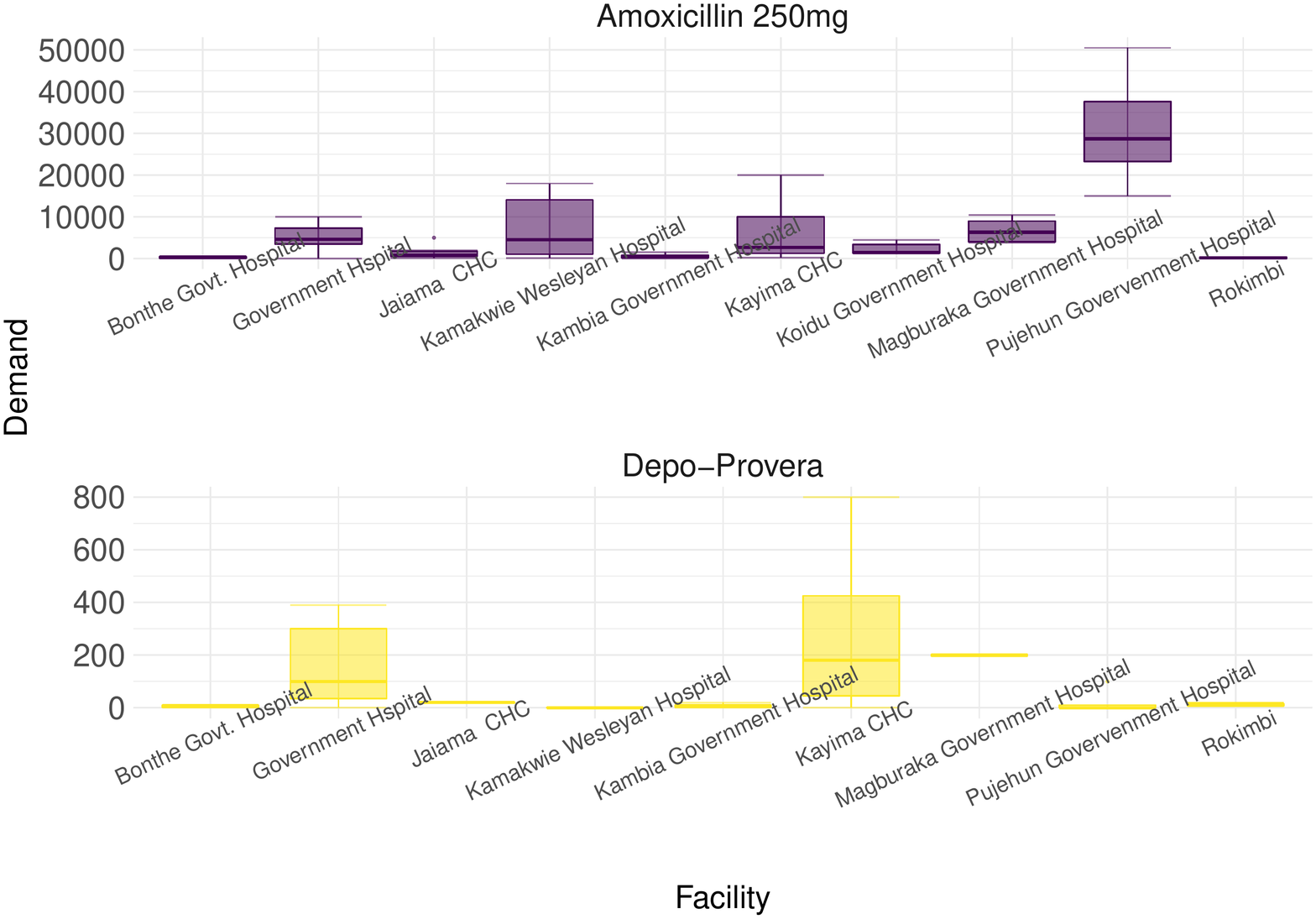}
\caption{Demand distribution of selected facility-product pairs from 2021-2022}
\label{fig:dist}
\end{figure}

\textbf{Distribution Pattern.} One of the challenges to make optimal prediction and allocation is highly uncertain demand in Sierra Leone. Fig.~\ref{fig:dist} shows that the demand of each facility-product pair fluctuates greatly over time, and the degree of fluctuations also differ greatly across facilities and products.

\textbf{Features.} Applying time series prediction methods fared poorly in our setting because we have too many time series (one for each $1040\times 8$ facility-product pair, for essential medicines) and each time series has too few observations (at most 34, but on average, 40\% of the observations per facility were missing/unreliable). Thus, we face a very high-dimensional, small data problem. Therefore, as described earlier, we use a multitask learning approach, learning simultaneously across facilities and products. We also performed extensive feature engineering based on domain knowledge: we constructed features including historical demand for the past 10 months, the current month and year, the region of the facility, etc., for a total of 40 features for each facility (i.e., $x_n\in\mathbb{R}^{40}$, and $x\in\mathbb{R}^{1040\times40}$). 

\begin{figure}[h]
\begin{center}
\includegraphics[width=0.8\columnwidth]{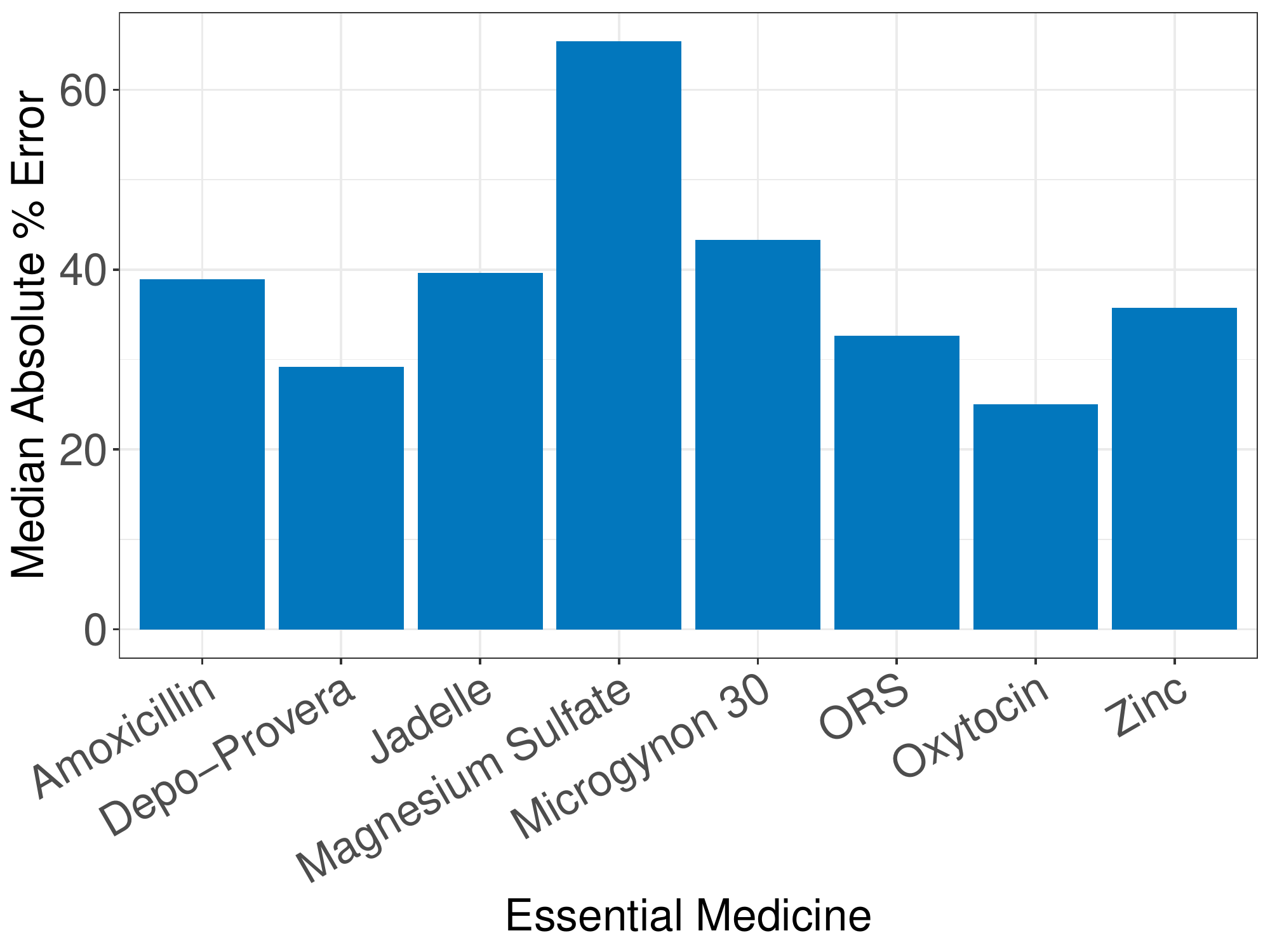}
\caption{Median Absolute Percentage Error (\%) of decision-blind random forest model.}
\label{fig:MAPE}
\end{center}
\end{figure}

\textbf{Prediction.} The resulting median absolute percentage error (MdAPE) for the (decision-blind) random forest is shown in Fig.~\ref{fig:MAPE}. We observe that despite the noisy data, we are able to predict demand reasonably well. Note that our decision-aware approach retrains this random forest based on sample weights given by the decision loss.

\begin{figure}[h]
\begin{center}
\includegraphics[width=0.9\columnwidth]{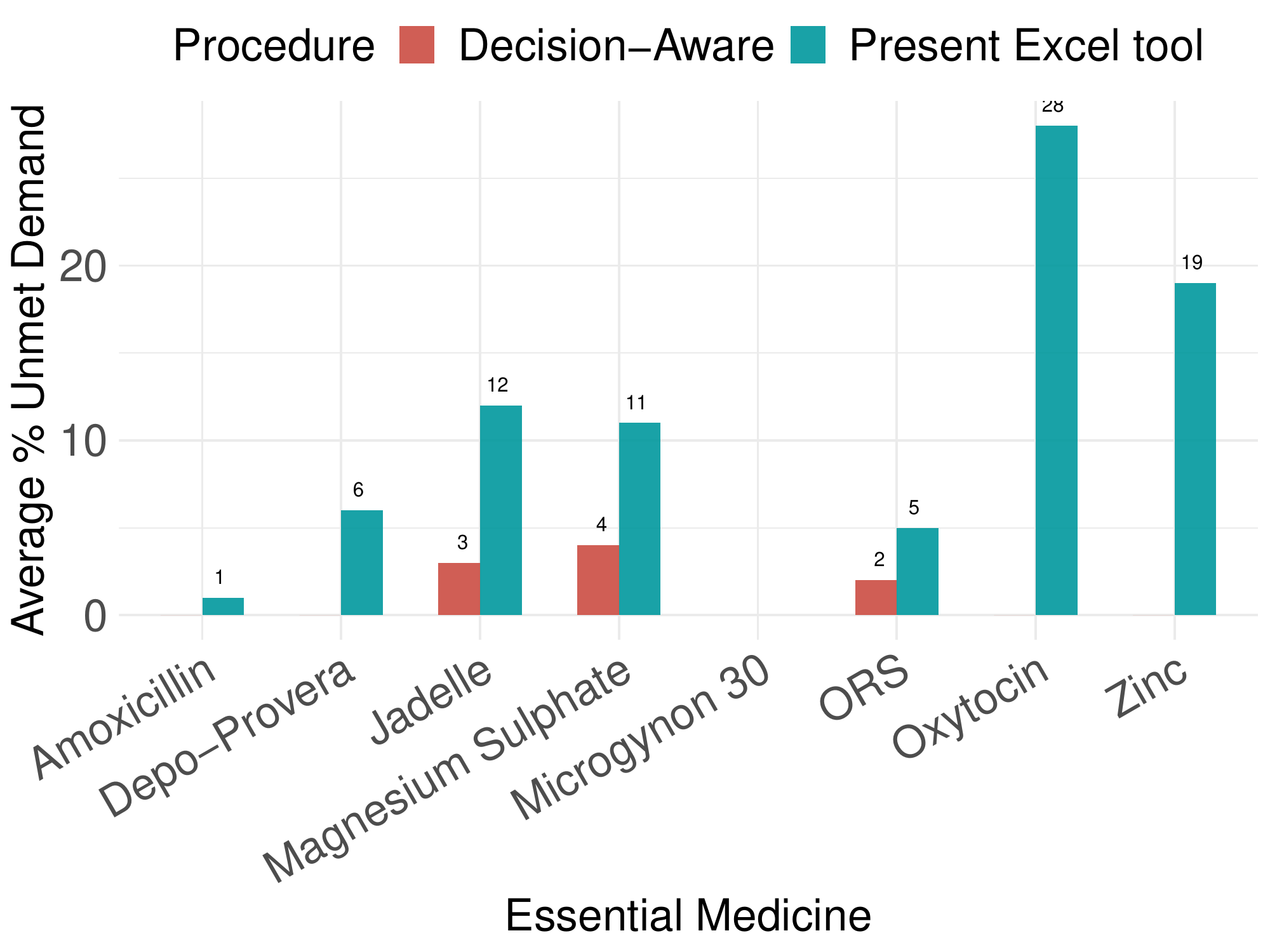}
\caption{Unmet demand comparison between existing practice and our decision-aware approach}
\label{fig:CPtool}
\end{center}
\end{figure}

\textbf{Comparison with Existing Practice.} We compared our approach to the existing strategy used by Sierra Leone. They currently use a complex Excel tool to decide the allocation mainly based on a three month rolling average of demand. We find that our approach has only average 1.3\% of unmet demand across eight products while the current tool is 15\% as shown in Fig.~\ref{fig:CPtool}.  Thus, our approach reduces unmet demand by an order of magnitude.

\end{document}